\documentclass[letterpaper, 10 pt, conference]{ieeeconf}

\IEEEoverridecommandlockouts                              
\usepackage{graphicx}
\usepackage{siunitx}
\usepackage[utf8]{inputenc}
\usepackage{url}
\usepackage[T1]{fontenc}
\usepackage{amsfonts}
\newcommand{\fig}[1]{Fig.~\ref{#1}}
\newcommand{\tab}[1]{Table~\ref{#1}}

\usepackage{epsfig} 
\usepackage{amsmath} 
\usepackage{amssymb}  
\usepackage{subcaption}
\usepackage{caption}
\usepackage{color}
\usepackage{verbatim}
\usepackage{wasysym}
\usepackage[table,xcdraw]{xcolor}
\usepackage{graphicx}
\usepackage{gensymb}
\usepackage{xcolor}
\usepackage{latexsym}
\usepackage{multirow}
\usepackage{amsmath}
\usepackage{booktabs,caption}
\usepackage[noadjust]{cite}
\usepackage[flushleft]{threeparttable}
\usepackage[table,xcdraw]{xcolor}
\usepackage{tabularx}

\usepackage[normalem]{ulem} 
\usepackage{wasysym}

\newcommand{\RNum}[1]{\uppercase\expandafter{\romannumeral #1\relax}}
\makeatletter
\newlength\tmp@\newlength\t@mp
\newcommand{\comp}[3]
  {\mathop{ \settowidth\tmp@{$\displaystyle\mathop{#1}^{#3}_{#2}$}
  \hbox to \tmp@{\hss \settowidth\t@mp{$\displaystyle #1$}\setlength\t@mp{.45\t@mp}
  $\displaystyle\mathop{#1}^{\hspace\t@mp #3}_{\hspace{-\t@mp}#2}$
  \hss} }}

\makeatother
\title{\LARGE \bf
Mechanical Intelligence-Aware Curriculum Reinforcement Learning \\ for Humanoids with Parallel Actuation\\
}

\author{
Yusuke Tanaka$^{1*}$, Alvin Zhu$^{1*}$, Quanyou Wang$^{1}$, Yeting Liu$^{1}$, Dennis Hong$^{1}$%
\thanks{$^{1}$Department of Mechanical and Aerospace Engineering, UCLA, Los Angeles, CA, USA. Emails: {\tt\small \{yusuketanaka, alvister88, dennishong\}@g.ucla.edu}. *Equal contribution.}%
}

\begin{document}


\maketitle

\begin{abstract}
Reinforcement learning (RL) has enabled advances in humanoid robot locomotion, yet most learning frameworks do not account for mechanical intelligence embedded in parallel actuation mechanisms due to limitations in simulator support for closed kinematic chains. 
This omission can lead to inaccurate motion modeling and suboptimal policies, particularly for robots with high actuation complexity. 
This paper presents general formulations and simulation methods for three types of parallel mechanisms: a differential pulley, a five-bar linkage, and a four-bar linkage, and trains a parallel-mechanism aware policy through an end-to-end curriculum RL framework for BRUCE, a kid-sized humanoid robot. Unlike prior approaches that rely on simplified serial approximations, we simulate all closed-chain constraints natively using GPU-accelerated MuJoCo (MJX), preserving the hardware's mechanical nonlinear properties during training. We benchmark our RL approach against a model predictive controller (MPC), demonstrating better surface generalization and performance in real-world zero-shot deployment. 
This work highlights the computational approaches and performance benefits of fully simulating parallel mechanisms in end-to-end learning pipelines for legged humanoids. Project codes with parallel mechanisms:
\url{https://github.com/alvister88/og_bruce}
\end{abstract}

\section{Introduction} 
Humanoid robots have achieved significant advancements in mobility and manipulation through various control strategies, particularly Model Predictive Control (MPC) \cite{MIT_humanoid, bruce} and Reinforcement Learning (RL) \cite{digit_science, deepmind_soccer}. However, purely learning-based approaches have yet to leverage the mechanical intelligence embedded in robot designs fully \cite{parallel_simulation}. In particular, parallel mechanisms have been shown to offer advantages such as higher combined motor output power, reduced inertia, and greater transmission ratio \cite{five_bar_leg, shin2019_five_bar}. 
Despite benefits, the parallel mechanisms are less commonly modeled in robotics physics simulations due to the challenges of handling closed kinematic chains, where a single child body is connected to more than one parent body \cite{digit_science}. 

Existing simulation frameworks often sidestep the complexity of closed-chain systems by simplifying or approximating the parallel linkages as serial chains \cite{urdf_plus}. Such assumptions undermine the natural mechanical intelligence inherent in the structure, requiring a lower-level controller that handles the actual parallel mechanisms \cite{parallel_simulation}. Consequently, controllers trained or designed under these simplified conditions cannot fully exploit the system’s real-world mechanical properties.
Furthermore, these approximations can lead to inaccurate and suboptimal motion \cite{parallel_simulation} in the simulation, as well as an inability to simulate singularities, which is a notable issue in parallel mechanisms \cite{yusuke_scaler_2022}. 

\begin{figure}
    \centering
    \includegraphics[width=0.7\linewidth,trim={0cm 0cm 0cm 0cm},clip]{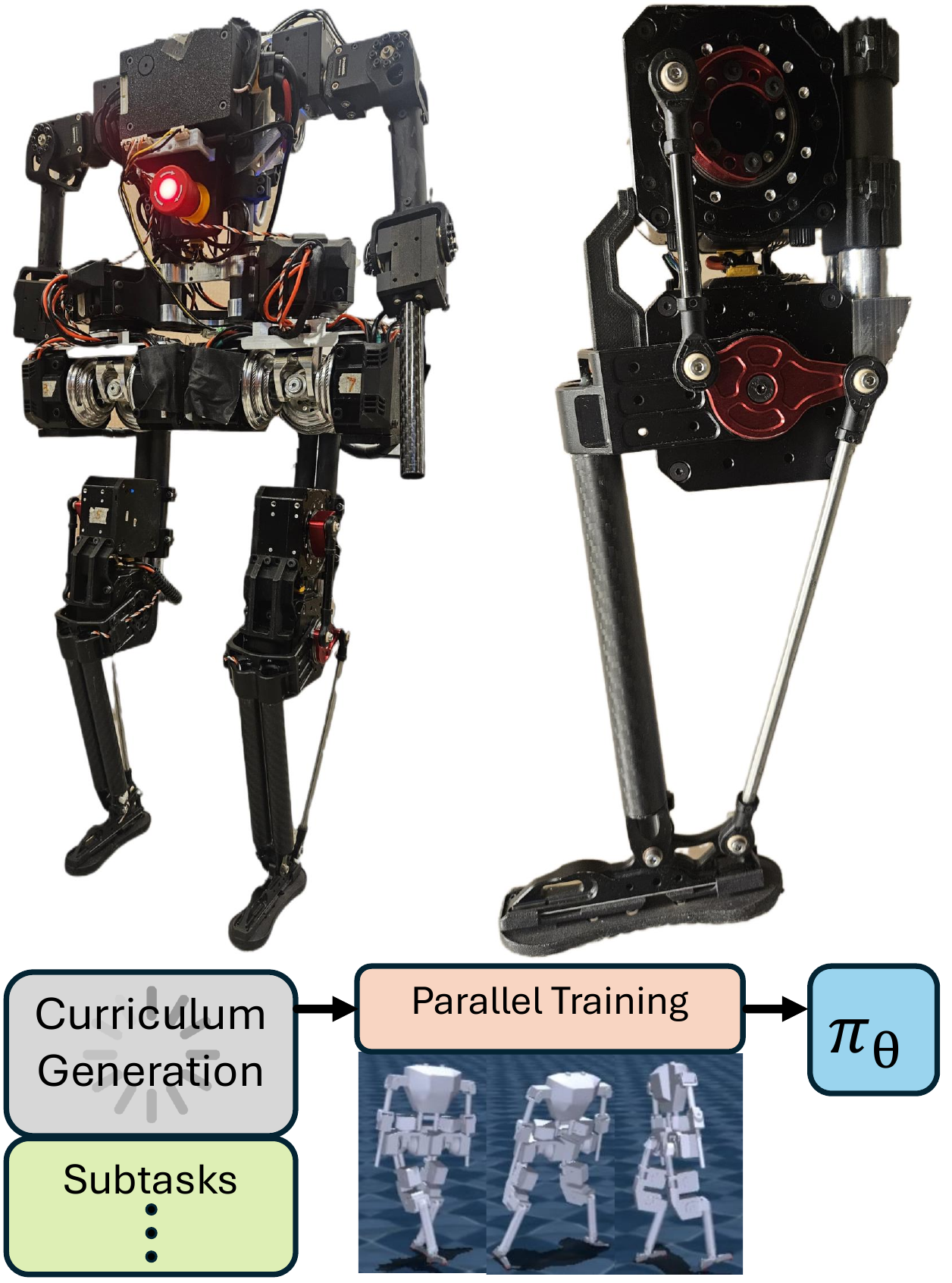}
    \caption{BRUCE \cite{bruce} hardware with three distinct parallel mechanisms, which are simulated, and an RL policy is deployed on the hardware zero-shot.}
    \label{fig:fig1}
\end{figure}


This paper presents general formulations of three parallel mechanisms for GPU-accelerated simulation in MuJoCo (MJX), leveraging soft equality constraints to preserve the inherent benefits of parallel hardware. We train locomotion policies via curriculum reinforcement learning directly on these fully simulated parallel chains and deploy them on BRUCE \cite{bruce}, a kid-sized humanoid robot.
The closed-chain formulation preserves actuator-to-output mappings for position, velocity, and torque, while explicitly representing four- and five-bar singularities.
This enables an end-to-end control policy in which the action space maps directly to the physical hardware actuators, removing the need for forward or inverse kinematics. An overview of the BRUCE hardware is provided in \fig{fig:fig1}.
Finally, we compare the performance of these learned policies against an MPC controller, analyzing the trade-offs of incorporating parallel linkage modeling in both simulation and hardware deployment.

The main contributions of this work are: 
\begin{itemize} 
\item Simulating three distinct parallel mechanisms in the simulation to account for closed-chain, faithfully capturing actuator-level position and torque mappings and kinematic singularities.
\item A curriculum RL trained a fully end-to-end locomotion policy on a humanoid robot, BRUCE.
\item A hardware validation and performance comparison against a baseline MPC. 
\end{itemize}

By integrating parallel mechanism modeling directly into the RL training process and evaluating the results in hardware, this study highlights the importance and feasibility of capturing the full mechanical dynamics of complex humanoid robots, providing a way to embrace mechanical intelligence in machine learning.


\section{Related Work} 
\subsection{Parallel Mechanism Leg and Humanoid}
Parallel linkage mechanisms have been widely adopted in legged robotics thanks to their mechanical advantages, including transmission and structural benefits and combined actuator outputs \cite{five_bar_leg}.
These advantages have been leveraged in agile dynamic \cite{five_bar_leg, minitaur, delta_hopper} and power-intensive domains \cite{bobcat, yusuke_scaler_2022}, enabling novel locomotion performance.

In humanoid robot leg designs, parallel linkages reduce leg inertia and aggregate mass closer to the torso, approaching the idealized single-mass spring-loaded inverted pendulum model \cite{hubicki2016atrias, shin2019_five_bar}.
Various research humanoid legged robots have employed hybrid serial-parallel linkages \cite{disney_parallel_leg} to realize higher DoF in the leg \cite{kangaroo_humanoid, scaler-b} or through tendon-driven mechanisms \cite{kawaharazuka2023design}. 
Commercially developed humanoids (e.g., Unitree H1, Fourier GR1, and Optimus \cite{parallel_simulation}) also utilize parallel linkages, yet simulating such closed-chain mechanisms remains challenging.

Those parallel mechanism-based humanoid platforms have been successful using both model-based \cite{MIT_humanoid} and RL \cite{digit_science} methods. However, simulating the closed kinematics chain has been challenging \cite{urdf_plus, makabe2024design}.
The GPU-accelerated Isaac Gym RL framework lacks native support for closed-chain kinematics, requiring custom implementations and approximations \cite{digit_science}. 
Researchers have resorted to strategies such as adopting a kinematic rather than actuator joint space or approximating the parallel mechanism as a serial chain \cite{parallel_simulation}.
These approaches can yield suboptimal results \cite{parallel_simulation} and limit direct use of mechanical advantages.
Efforts to accurately simulate parallel linkages include extending URDF with graph-based pre-processing and constraint embedding \cite{urdf_plus} or treating closed chains as contact constraints with full differentiation for optimal control \cite{parallel_simulation}.

\subsection{Reinforcement Learning on Humanoid}
RL has emerged as an effective framework for robots to acquire complex control policies directly through interaction, without the need for explicit system modeling. Deep RL algorithms—such as Proximal Policy Optimization (PPO) \cite{OG_PPO}, Trust Region Policy Optimization (TRPO) \cite{trpo}, and Deep Deterministic Policy Gradient (DDPG) \cite{ddpg}—have demonstrated impressive performance on high-dimensional locomotion tasks in both bipedal and quadrupedal robots \cite{dog_trpo}. These algorithms enable agents to learn robust and adaptive behaviors by optimizing policies over time through experience, often in the presence of non-linear dynamics, contact-rich environments, and unstructured terrain \cite{robust_walking}.

However, the application of RL to robotics presents significant challenges \cite{rl_challenges}. These methods typically require large-scale data collection and finely tuned reward functions to ensure stable learning \cite{large_data_rl}. In humanoid robotics, this is especially difficult due to the system’s high degrees of freedom (DoF), underactuation, and sensitivity to small control errors \cite{learn_walking}. To bridge the gap between simulation and the real world, sim-to-real transfer has become an essential focus \cite{simtoreal1, simtoreal2}. Techniques such as domain randomization, dynamics perturbation, and injecting latency into the environment are commonly employed to expose policies to a wide range of conditions during training, improving their robustness to real-world discrepancies \cite{domain_randomization}. Recent advancements also explore zero-shot policy transfer, where agents trained entirely in simulation can generalize to real-world deployment through domain randomization \cite{BALLU_RL} without additional fine-tuning \cite{HumanoidTransformer, manip_zero_shot}. Curriculum learning is also often used to improve convergence by structuring training from simple to more complex tasks, helping guide policy development progressively and stably \cite{curriculum1, reverse_curriculum, zhu2025aura}.

We present formulations of three parallel mechanisms using soft equality constraints, and demonstrate curriculum RL-trained locomotion policies on fully simulated parallel chains, deployed on BRUCE.

\section{Parallel Mechanism and Closed Kinematic Chain\label{sec:parallel}}
\begin{figure}
    \centering
    \includegraphics[width=0.99\linewidth,trim={0cm 0cm 0cm 0cm},clip]{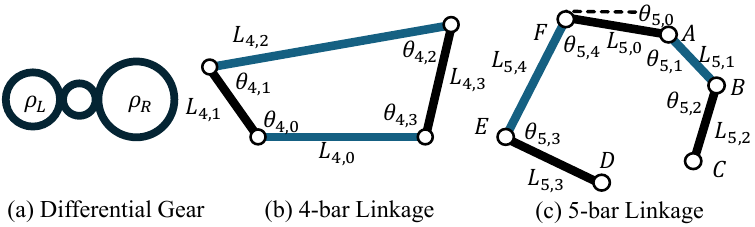}
    \caption{Generic topology definitions of the BRUCE's parallel mechanisms. 
    }
    \label{fig:linkage_def}
\end{figure}

This section covers the general mathematical formulation of the three closed kinematic chain mechanism constraints. 
BRUCE \cite{bruce} is an agile and open-source small-sized humanoid platform. 
BRUCE includes three distinct parallel linkages, making it challenging to simulate the mechanisms entirely. The open-source simulation uses the kinematic joint space. Their general topologies are defined in \fig{fig:linkage_def}.

\begin{figure}
    \centering
    \includegraphics[width=0.99\linewidth,trim={3cm 0cm 0cm 0cm},clip]{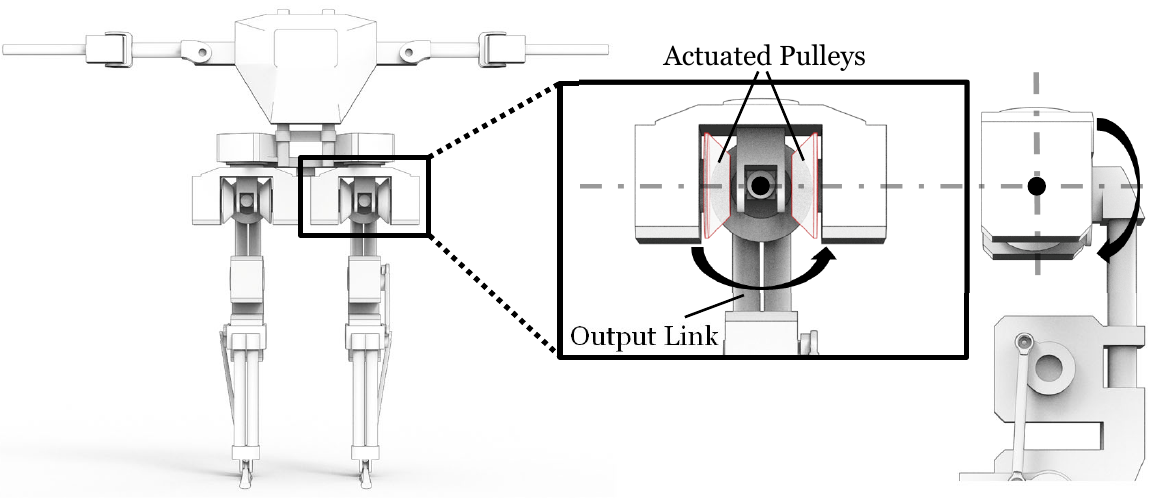}
    \caption{Close-up of BRUCE’s cable-driven differential pulley.}
    \label{fig:bruce_diff}
\end{figure}

\subsection{Hip Differential Drive Gear Mechanism\label{sec:differential}} 
BRUCE's hip joint employs a cable-driven differential pulley system \cite{bruce}, which forms a 2-DoF parallel mechanism with minimal backlash.
This differential mechanism as shown in \fig{fig:bruce_diff} imposes equality constraints between the actuated input and passive output joints, which can be mathematically expressed as:
\begin{equation}
\begin{bmatrix}
\dot{q}_{\text{hip, roll}} \\
\dot{q}_{\text{hip, pitch}}
\end{bmatrix}
=
\frac{1}{\rho_L + \rho_R}
\begin{bmatrix}
\rho_L & \rho_R \\
\rho_L & -\rho_R
\end{bmatrix}
\begin{bmatrix}
\dot{q}_L \\
\dot{q}_R
\end{bmatrix}
\end{equation}

Here, $\dot{q}_{\text{hip, roll}}$ and $\dot{q}_{\text{hip, pitch}}$ denote the roll and pitch joint velocities, respectively.  
$\dot{q}_{L,R}$ denote the actuator velocities, and $\rho_{L,R}$ represent the gear ratios between the left/right drive bevel gears and the central idler gear.

\begin{figure}
    \centering
    \includegraphics[width=0.9\linewidth,trim={0cm 0cm 0cm 0cm},clip]{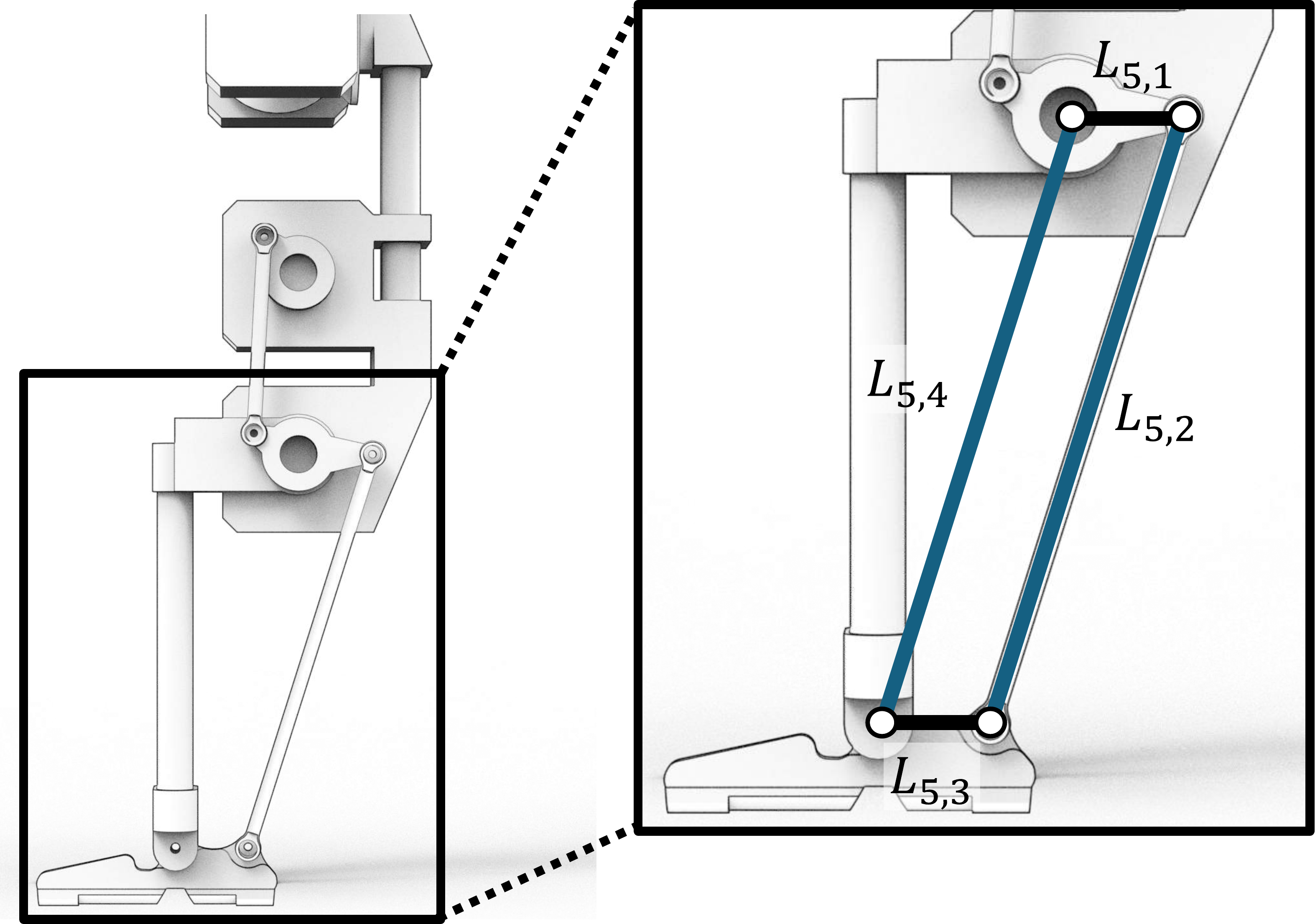}
    \caption{Close-up of BRUCE’s five-bar parallelogram linkage. $L_{5,1}$ and $L_{5,4}$ are actuated by two motors. $L_{5,0}$ in \fig{fig:linkage_def} is zero.}
    \label{fig:bruce_5bar}
\end{figure}

\subsection{Leg Five-Bar Linkage Mechanism\label{sec:5-bar}}
The kinematic simulation of the five-bar linkage shown in \fig{fig:bruce_5bar} is modeled as two serial chains whose endpoints are constrained to coincide, forming a closed loop.
The first endpoint \(C\) and the second endpoint \(D\) is given as:
\begin{align}
\mathbf{p}_C &= \mathbf{R}(\theta_{5,1})
\begin{bmatrix} l_{5,1} & 0 \end{bmatrix}^{\!\top}
+\mathbf{R}(\theta_{5,2})
\begin{bmatrix} l_{5,2} & 0 \end{bmatrix}^{\!\top},\\
\mathbf{p}_D &= \mathbf{R}(\theta_{5,4})
\begin{bmatrix} l_{5,4} & 0 \end{bmatrix}^{\!\top}
+\mathbf{R}(\theta_{5,3})
\begin{bmatrix} l_{5,3} & 0 \end{bmatrix}^{\!\top}.
\end{align}
Note that the rotation matrix is $\mathbf{R}(\theta)\in \mathrm{SO}(2)$. 
These points are projected into 3D space:
\begin{equation}
\tilde{\mathbf{p}}_C = {}^{\text{b}}T_A 
\begin{bmatrix} \mathbf{p}_C & 0 & 1 \end{bmatrix}^\top, \quad
\tilde{\mathbf{p}}_D = {}^{\text{b}}T_F 
\begin{bmatrix} \mathbf{p}_D & 0 & 1 \end{bmatrix}^\top
\end{equation}

Where the transformation matrices \(^{\text{b}}T_{A}\) and \(^{\text{b}}T_{F}\) represent the 3D poses of points \(A\) and \(F\) relative to the base frame. 
The closed-chain constraint is then enforced as $\tilde{\mathbf{p}}_C \approx \tilde{\mathbf{p}}_D.$

\(\theta_{5,1}\) and \(\theta_{5,4}\) are actuated joints, while \(\theta_{5,2}\) and \(\theta_{5,3}\) are passive.

\subsubsection{Handling Multiple Solutions}
The 5-bar linkage can have five possible kinematic solutions given an endpoint position. Hence, starting the linkages with feasible and desirable configurations is necessary. If the singularity of the mechanisms is a concern in the simulation, a sanity check during the simulation is recommended.

\begin{figure}
    \centering
    \includegraphics[width=0.8\linewidth,trim={0cm 0cm 0cm 0cm},clip]{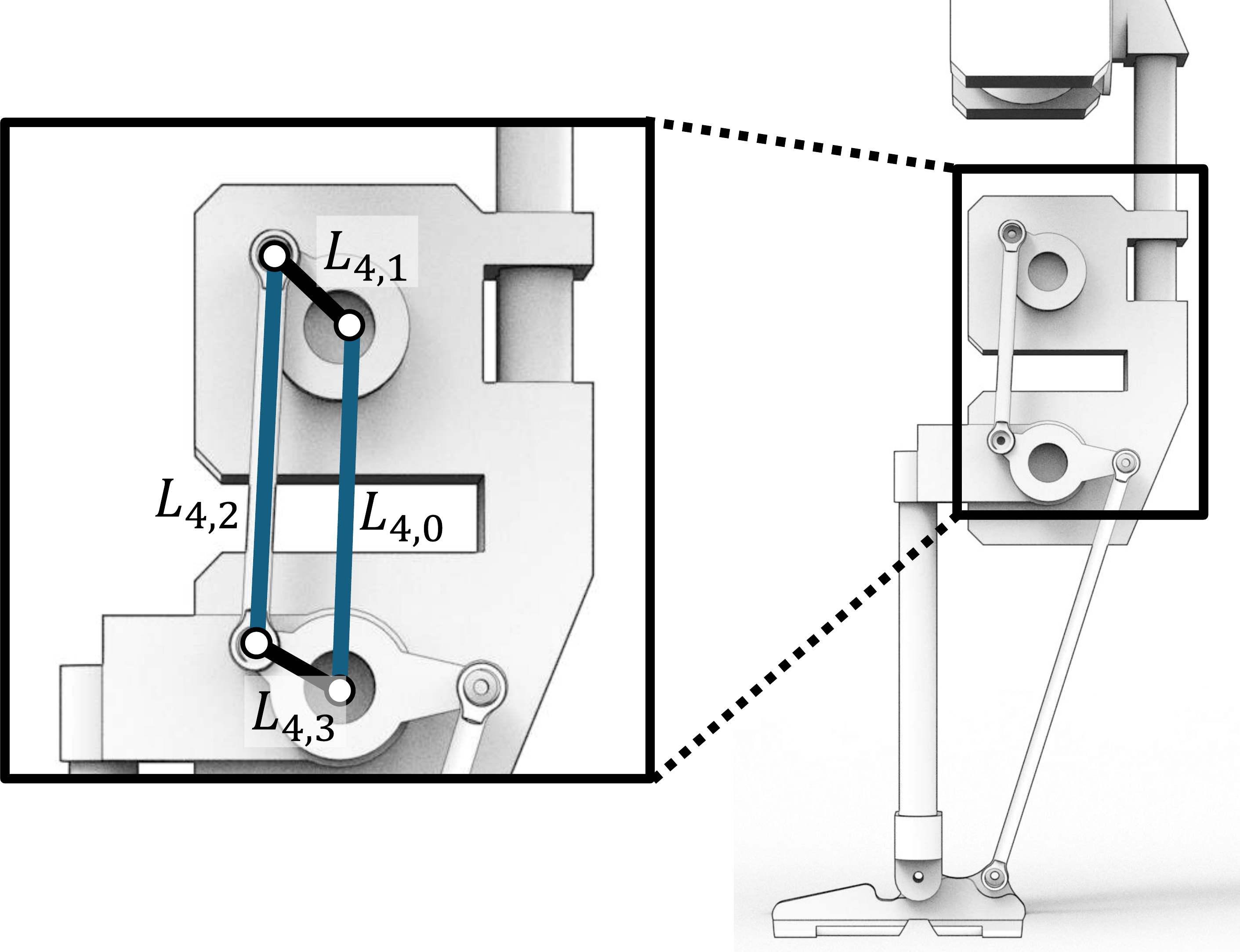}
    \caption{Close-up of BRUCE’s four-bar linkage. $L_{4,1}$ is actuated.}
    \label{fig:bruce_4bar}
\end{figure}

\subsection{Four-Bar Linkage Mechanism}
The four-bar linkage shown in \fig{fig:bruce_4bar} is a special case compared to five-bar mechanisms or differential gears, as it has only one actuator and one output DoF.
Thus, it can be viewed similarly to a gear with a nonlinear transmission ratio, which can be approximated using a polynomial function. However, unlike gears, the 4-bar linkage may encounter singularities (e.g., when all the links are collinear), which cannot be accurately captured by polynomial approximation. 
Therefore, we propose two approaches for modeling the 4-bar linkage.

\subsubsection{Polynomial Transmission Ratio Approximation\label{sec:4_bar_method1}}
The 4-bar linkage acts as a mechanism that alters the center of rotation, with a transmission ratio determined by the lengths.
\begin{equation}
\rho_4 =\frac{\dot{\theta}_{4,0}}{\dot{\theta}_{4,3}} =
\frac{L_{4,2} \sin(\theta_{4,1} - \theta_{4,2})}
     {L_{4,2} \sin(\theta_{4,1} - \theta_{4,2}) - L_{4,3} \sin(\theta_{4,1} - \theta_{4,3})}
\label{eq:angular_velocity_ratio}
\end{equation}

Here, $\rho_4$ denotes the velocity transmission ratio between the input angle $\theta_0$ and the output angle $\theta_3$. 
The torque transmission ratio is the reciprocal of $\rho$. In the special case where the 4-bar forms a parallelogram, this ratio becomes constant: $\rho = l_1 / l_3$.

To approximate the nonlinear mapping between input and output angles, we use a polynomial constraint of the form:
\begin{align}
\mathbf{r}_{f} &= y - y_0 - \mathbf{a}^T \boldsymbol{\phi}(x - x_0) = 0 \\
\mathbf{a} &= \begin{bmatrix} a_0 & \cdots & a_i \end{bmatrix}^T, i \in \mathbb{N} \\
\boldsymbol{\phi}(x - x_0) &= \begin{bmatrix} (x - x_0)^0 & \cdots & (x - x_0)^i \end{bmatrix}^T
\end{align}
Here, $x_0$ and $y_0$ denote the nominal configuration point for the input and output angles, respectively.

The coefficient vector $\mathbf{a}$ can be obtained via least squares regression to fit the observed mapping between $x = \theta_{4,0}$ and $y = \theta_{4,3}$ over a feasible range of motion, as discussed in Section~\ref{sec:4_bar_method2}.

\subsubsection{Closed loop kinematic link constrained representation\label{sec:4_bar_method2}}
While the method in Section~\ref{sec:4_bar_method1} can capture the position–torque relationship of the four-bar linkage, it does not model singularities—particularly when all four links become collinear.
This limitation can be mitigated if the mechanism physically prevents reaching such configurations via joint limits. Otherwise, a constraint representation similar to the five-bar linkage in Section~\ref{sec:5-bar} is necessary. 


\begin{figure*}[t]
    \centering
    \includegraphics[width=0.95\linewidth]{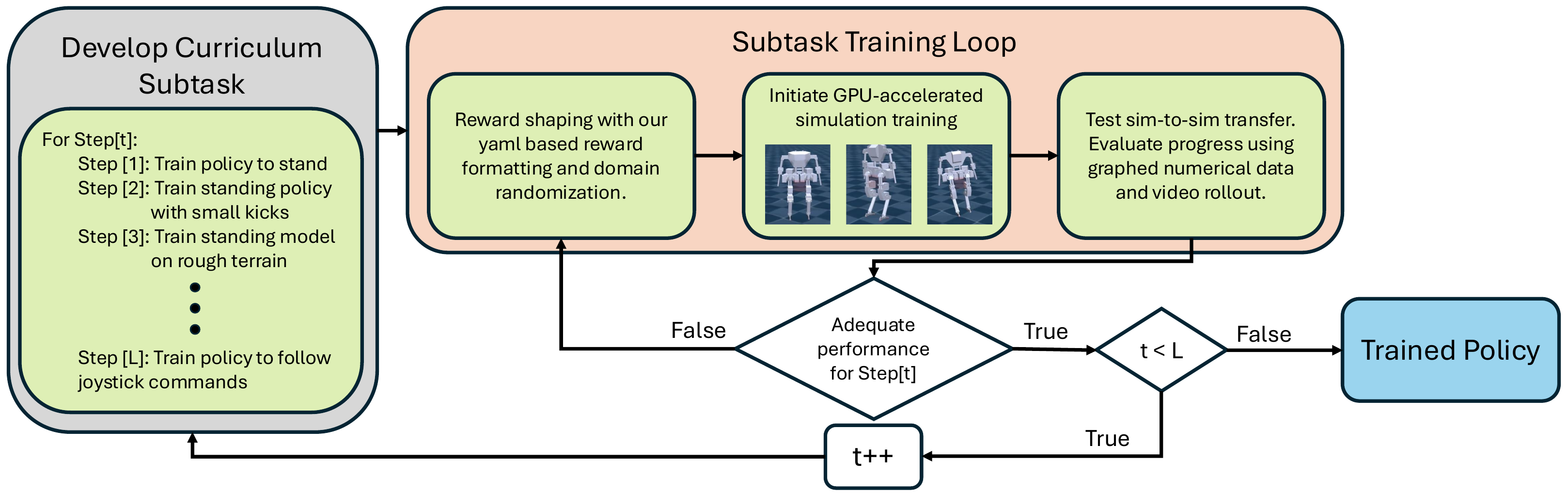}
    \caption{Curriculum reinforcement learning framework overview.}
    \label{fig:curriculum}
\end{figure*}

\subsection{Backlash Due to Passive Joints}

One mechanical drawback of parallel mechanisms is the introduction of unintended compliance, particularly at passive joints that are indirectly driven by actuated ones. Such backlash can be captured in a simulation when the full kinematics of the parallel mechanism are explicitly modeled.

In MuJoCo \cite{todorov2012mujoco}, all equality constraints are treated as soft and enforced through a virtual spring-damper system modulated by a constraint impedance function:
\begin{equation}\label{eq:mujoco_const}
    a_{c1} + \mathcal{D}(r) \cdot (b_v v + k_v r) = (1 - \mathcal{D}(r)) \cdot a_{c0}
\end{equation}

Here, $a_{c1}$ is the constrained acceleration due to the applied constraint force, and $a_{c0}$ is the unconstrained acceleration. The function $\mathcal{D}(r)$ is a smooth, symmetric polynomial sigmoid that modulates the constraint impedance as a function of the constraint violation $r$. The coefficients $b_v$ and $k_v$ represent damping and stiffness, respectively, and $v$ is the relative velocity at the constraint.

To model backlash, $\mathcal{D}(r)$ is designed to remain low within a deadband region (i.e., $|r| < \epsilon_q$), allowing low impedance motion. As $|r|$ exceeds this threshold, $\mathcal{D}(r)$ increases smoothly toward its maximum, gradually enforcing the constraint. The sigmoid is governed by hyperparameters defining the inflection point and steepness of the sigmoid.

\section{Reinforcement Learning}

\begin{figure}
    \centering
    \includegraphics[width=0.9\linewidth]{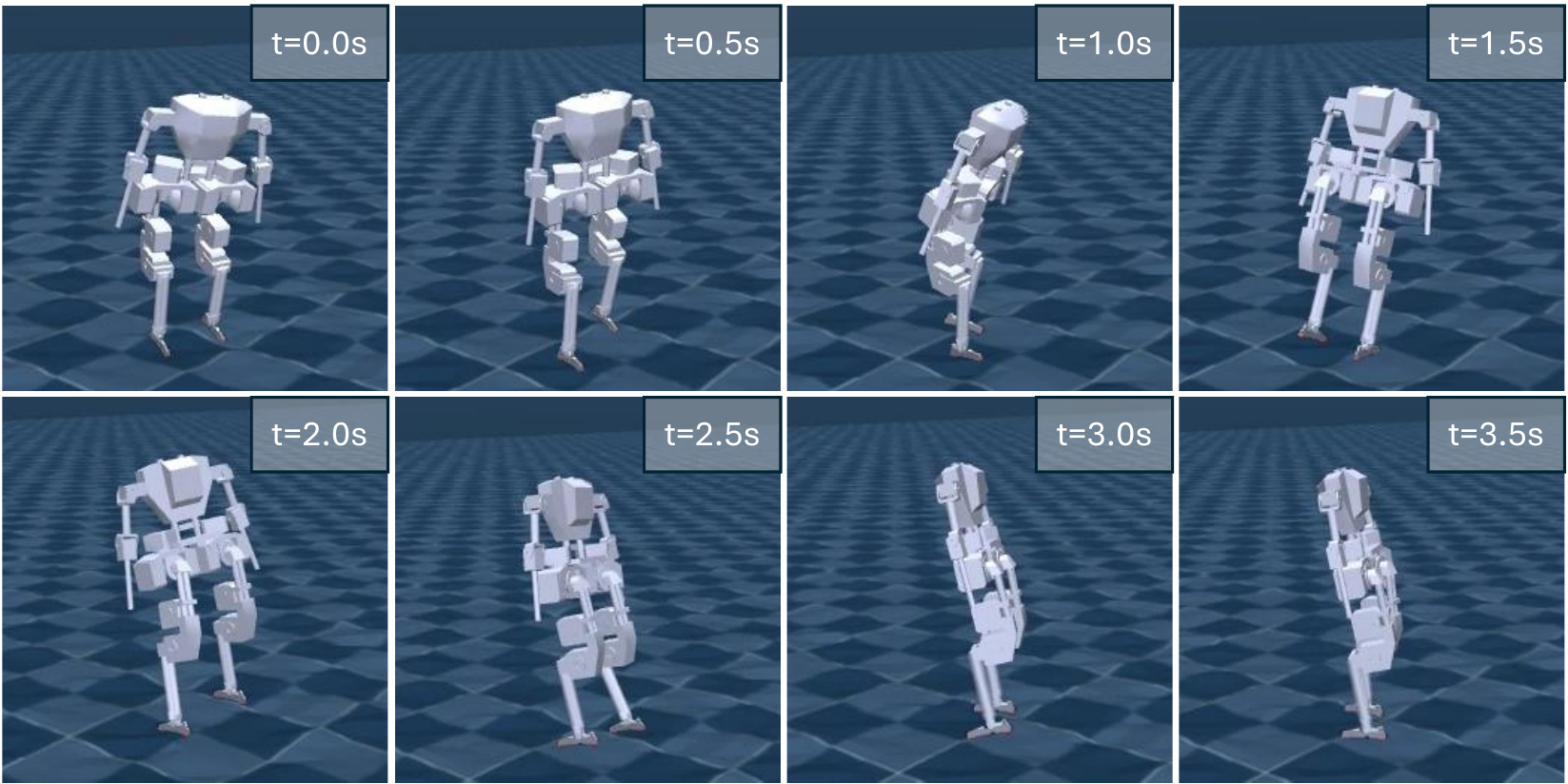}
    \caption{BRUCE RL policy in MuJoCo sim-to-sim transfer. MuJoCo and MJX use different physics backends.}
    \label{fig:mujoco_siml}
\end{figure}

\subsection{RL Formulation}
We model the humanoid locomotion task as a Markov Decision Process \cite{mdp} defined by the tuple $(\mathcal{S}, \mathcal{A}, P, r, \gamma)$, where $\mathcal{S}$ is the state space, $\mathcal{A}$ the action space, $P(s_{t+1} \mid s_t, a_t)$ the state transition probability, $r: \mathcal{S} \times \mathcal{A} \to \mathbb{R}$ the reward function, and $\gamma \in (0,1]$ the discount factor.

\subsection{Action Space}
In this paper, we explore three different action spaces: positional, residual position command, and residual position command with respect to the previous reference. As discussed in Section~\ref{sec:parallel}, all of our actuated joints in the simulation are the same as the hardware actuator joints.  
\textbf{Action Space:}  
The action $a_t \in \mathbb{R}^{16}$ at time $t$ is represented by a vector, which encodes the positional commands for each of the actuator joints of the humanoid robot. These commands are interpreted as offsets relative to a nominal home configuration $q_{\text{nom}} \in \mathbb{R}^{16}$, such that the reference joint angles are given by 
$q_t = q_{\text{nom}} + a_t$.

\subsubsection{Observation Space}  
To capture temporal dependencies and mitigate sensor noise and latency, we maintain a history of $H$ observations. The aggregated vector, $\mathbf{O}_t$, is defined as:
\begin{align}\label{eq:obs}
    \mathbf{O}_t = \begin{bmatrix}
    \mathbf{o}_t & \mathbf{o}_{t-1}& \cdots & \mathbf{o}_{t-(H-1)}
    \end{bmatrix}^T, \quad \mathbf{O}_t \in \mathbb{R}^{N_o \cdot H} \\
\mathbf{o}_{t} = \Bigl[ \dot{\psi}_{\text{IMU},t} \quad \mathbf{g}_{\text{proj},t} \quad \mathcal{C}_t \quad
\mathbf{q}_{t} - \mathbf{q}_{\text{nom}} \quad \mathbf{a}_{t-1} \Bigr]
\end{align}

$\mathbf{o}_t$ contains the observation at timestep $t$, such as sensor readings \eqref{eq:obs}.
 $\dot{\psi}_{\text{IMU}}$ is the yaw rate measured by IMU, $\mathbf{g}_{\text{proj}}$ is a projected gravity vector representing the robot base frame tilt with respect to the world gravity vector, and $\mathcal{C}_t = (c_x, c_y, c_{\omega_\psi})$ are the user velocity commands in $x$, $y$ and yaw.  $\mathbf{q_t} - \mathbf{q}_{\text{nom}}$ is the difference between the current measured joint angle and the nominal positions, and $\mathbf{a}_{t-1}$ is the action from the previous timestep. 

On hardware, the projected gravity vector in \eqref{eq:projected_g} is computed from the rotation matrix estimated by the Madgwick filter \cite{madgwick2010efficient} using IMU gyro and acceleration data, $\mathcal{R}_{\text{IMU}} \in \mathrm{SO}(3)$. The Madgwick filter provided a more stable estimation for our IMU sensor than the complementary filter.
\begin{equation}
    \mathbf{g}_{\text{proj}} = \mathcal{R_{\text{IMU}}}^{-1}(0, 0, -1)\label{eq:projected_g}
\end{equation}

\subsubsection{Policy and Objective}  
We parameterize the agent's policy by $\theta$, with the policy represented as a function:
$\pi_\theta: \mathbb{R}^{N_o \cdot H} \to \mathcal{P}(\mathcal{A})$
which maps the aggregated observation $\mathbf{O}_t$ to a probability distribution over actions. The objective is to maximize the expected cumulative reward:
\begin{equation}
    J(\theta) = \mathbb{E}_{\pi_\theta}\left[\sum_{t=0}^{T} \gamma^t\, r(s_t, a_t)\right].
\end{equation}

\subsubsection{Training via Proximal Policy Optimization (PPO)}  
We employ PPO to update the policy in a stable manner. Let
\[
r_t(\theta) = \frac{\pi_\theta(a_t|s_t)}{\pi_{\theta_{\text{old}}}(a_t|s_t)}
\]
denote the probability ratio and $A_t$ the advantage estimate at time $t$. The PPO objective is given by:
\[
L^{\text{CLIP}}(\theta) = \mathbb{E}_t \left[\min\left(r_t(\theta) A_t,\; \text{clip}\left(r_t(\theta), 1-\epsilon, 1+\epsilon\right) A_t\right)\right],
\]
where $\epsilon$ is a hyperparameter that constrains the policy update.

\textbf{Environment Dynamics and Filtering:}  
The environment is implemented using the GPU-accelerated physics engine MuJoCo (MJX) \cite{todorov2012mujoco} that simulates the robot dynamics with domain randomization, sensor noise, and latency effects. Specifically, the state transition is given by:
A second-order Butterworth lowpass filter and a deadband filter are applied to control signals and observations. These filters ensure that the control commands and sensor readings are smoothed and simulate the control latencies experienced on hardware.




\begin{table}[t]
\centering
\small
\setlength{\tabcolsep}{4pt}
\renewcommand{\arraystretch}{0.9}
\caption{Parameter specifications for final-stage balancing}
\label{tab:params}
\begin{tabularx}{\linewidth}{@{}lX@{}}
\toprule
\multicolumn{2}{c}{\textbf{Domain randomization}} \\ \midrule
Terrain              & Height map, $\Delta h\in[0,0.02]\,\mathrm{m}$. \\
Mass                 & $(1\pm0.2)\times$ nominal. \\
Actuator $K_p$       & Offset $\in[-20, 20]$. \\
Foot contact         & $(x,y,z)=(\pm0.006,\pm0.003,\pm0.003)\,\mathrm{m}$. \\
COM                  & Translation $\pm0.015\,\mathrm{m}$ (xyz). \\ \midrule

\multicolumn{2}{c}{\textbf{Disturbances}} \\ \midrule
Observation noise    & $\pm0.03$ (IMU \& joints). \\
Kicks                & Interval 50 steps, $v\in[0.2,0.45]\,\mathrm{m/s}$; small: 10 steps, $[0.05,0.1]\,\mathrm{m/s}$. \\
IMU perturbation     & Displacement $\pm0.006\,\mathrm{m}$; tilt $\pm0.06\,\mathrm{rad}$. \\ \midrule

\multicolumn{2}{c}{\textbf{Latency}} \\ \midrule
Observation          & $\mathcal{N}\!\big(0,10^2\,\mathrm{ms}^2\big)$ (20\,ms step). \\
Action               & $\mathrm{Uniform}\{0,1,2\}$ steps (0–40\,ms). \\
\bottomrule
\end{tabularx}
\end{table}

\subsection{Reward Design\label{sec:reward}}
The reward function comprises several components designed to encourage stable, efficient, and task-oriented locomotion. \textit{Tracking rewards} guide the agent to follow commanded linear and angular velocities, while penalizing vertical motion and rotational instability. \textit{Postural rewards}, such as those on torso orientation and angular velocity, promote balance and uprightness. \textit{Control penalties} discourage excessive torque use, abrupt action changes, and high-magnitude actions to ensure smooth and energy-efficient behavior. \textit{Gait-related terms}, including feet air time, foot slip penalties, and a feet phase reward, encourage natural stepping patterns, reliable ground contact, and correct timing of foot placement. Additional terms like \textit{stand\_still} and \textit{termination} ensure the robot maintains a nominal pose when idle and penalizes premature terminations, respectively. The environment randomization and disturbance parameters are summarized in Table~\ref{tab:params}, while Table~\ref{tab:reward_terms} lists each reward term. 

\begin{table}[t]
\centering
\caption{Reward Term Formulations for Final-Stage Locomotion}
\label{tab:reward_terms}
\begin{tabularx}{\linewidth}{@{}lX@{}}
\toprule
\multicolumn{2}{c}{\textbf{Tracking and Motion}} \\ \midrule
Tracking Linear Velocity     & $r = \exp\left(-\frac{\|\mathbf{v}_\text{cmd} - \mathbf{v}_\text{local}\|^2}{2\sigma^2}\right)$ \\
Tracking Angular Velocity    & $r = \exp\left(-\frac{(w_\text{cmd} - w_\text{base})^2}{2\sigma^2}\right)$ \\
Angular Velocity XY Penalty  & $r = -\|\boldsymbol{\omega}_{xy}\|^2$ \\
Orientation Penalty          & $r = -\| \text{rot\_up}_{xy} \|^2$ \\ \midrule

\multicolumn{2}{c}{\textbf{Control and Smoothness}} \\ \midrule
Torque Penalty               & $r = -\left( \| \boldsymbol{\tau} \|_2 + \| \boldsymbol{\tau} \|_1 \right)$ \\
Action Rate Penalty          & $r = -\| \mathbf{a}_t - \mathbf{a}_{t-1} \|^2$ \\ \midrule

\multicolumn{2}{c}{\textbf{Gait and Foot Contact}} \\ \midrule
Feet Air Time Reward         & $r = \mathbf{1}_{\|\mathbf{c}_\text{cmd}\| > \epsilon} \cdot \sum (t_\text{air} - t_\text{thresh}) \cdot \mathbf{1}_\text{contact}$ \\
Foot Slip Penalty            & $r = -\left( \| \mathbf{v}_\text{foot} \|_2 + \| \boldsymbol{\omega}_\text{foot} \|_2 \right) \cdot \mathbf{1}_\text{contact}$ \\
Feet Phase Reward            & $r = \mathbf{1}_{\|\mathbf{c}_\text{cmd}\| > \epsilon} \cdot \exp\left(-\frac{\| \mathbf{z}_\text{foot} - r_z \|^2}{2\sigma^2}\right)$ \\ \midrule

\multicolumn{2}{c}{\textbf{Posture and Termination}} \\ \midrule
Standstill Penalty          & $r = \mathbf{1}_{\|\mathbf{c}_\text{cmd}\| < \epsilon} \cdot \| \mathbf{q}_\text{joint} - \mathbf{q}_\text{default} \|_1$ \\
Early Termination Penalty    & $r = \mathbf{1}_{\text{done} \land (t < t_\text{max})} \cdot (-1.0)$ \\
\bottomrule
\end{tabularx}
\end{table}

\subsection{Training Scheme}
Our training pipeline follows a multi-stage curriculum that progressively increases task complexity, enhancing both policy robustness and efficiency. 
\fig{fig:curriculum} outlines the overall process, and \tab{tab:params} summarizes the parameter values. All the stages include domain randomization. Trained and sim-to-sim transferred policy is visualized in \fig{fig:mujoco_siml}.

The first stage trains the policy to maintain an upright posture. Rewards penalize deviations in the center of mass and body sway. 
In the second stage, small kicks are introduced to develop balance and recovery. The reward function penalizes imbalances while encouraging rapid stabilization.
The third stage introduces terrain irregularities, making it challenging to walk on uneven surfaces and randomizing the contacts. The reward structure includes penalties for slipping or falling.
In the final phase, the policy performs dynamic walking based on random commands. Rewards from previous stages are combined with task-oriented components.

The trained policy is evaluated to verify that the objective is achieved and that the policy is robust to various disturbances and domain randomization. This evaluation is done in both MJX and MuJoCo physics backends, validating the robustness through sim-to-sim transfer.

\section{Results and Hardware Experiments}

\begin{figure}[t]
    \centering
    \includegraphics[width=0.99\linewidth]{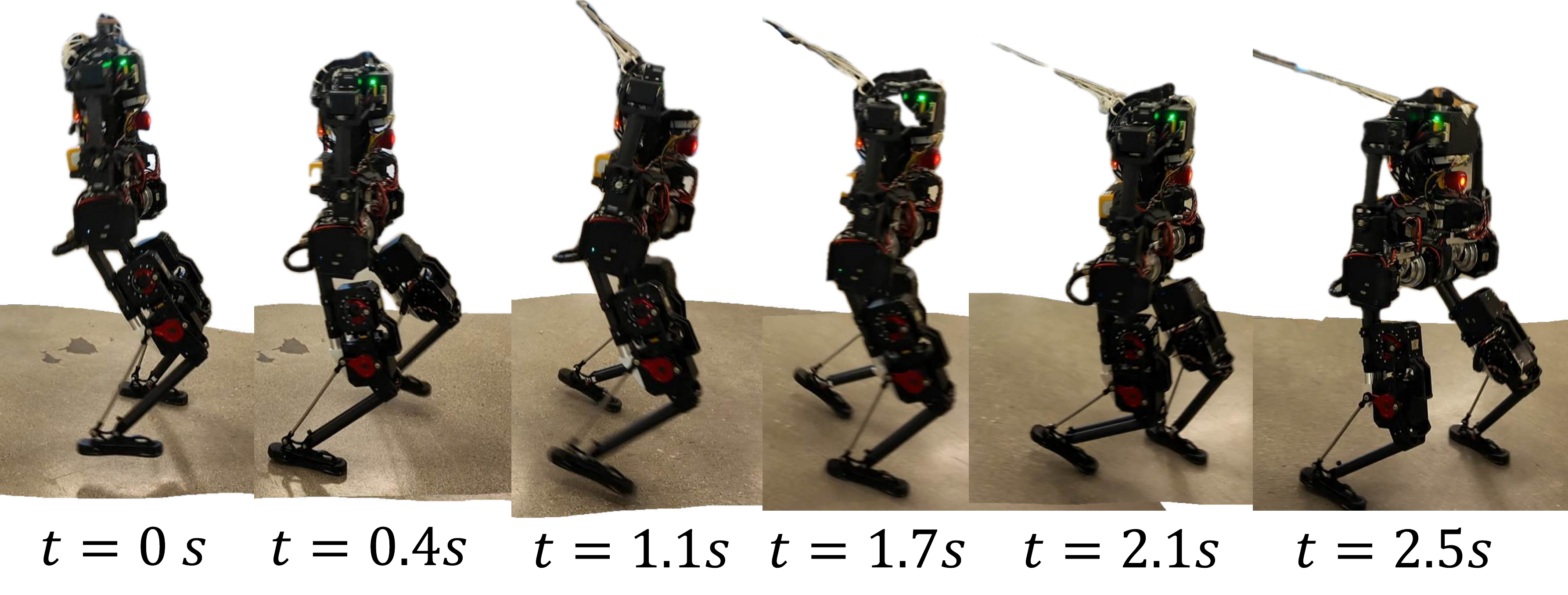}
    \caption{BRUCE RL policy locomotion on slippery smooth concrete surface. A safety leash is loosely attached at the top.}
    \label{fig:locomotion_smooth}
\end{figure}
\begin{figure}[t]
    \centering
    \includegraphics[width=0.99\linewidth]{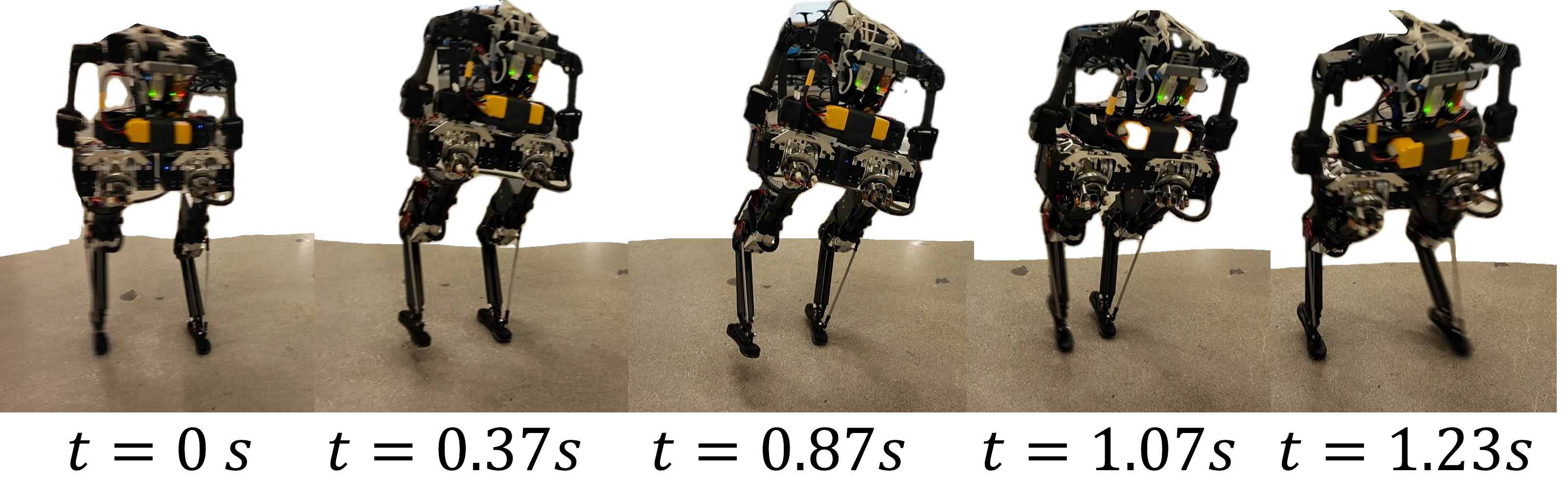}
    \caption{BRUCE RL policy standstill perturbation rejection in sideways. BRUCE (facing backward) was able to balance on only its right leg to stabilize itself, and then took one step back. After $t=\SI{0}{s}$ to $t=\SI{1.07}{s}$ the left leg was in the air. A safety leash is attached on top at loose. }
    \label{fig:one_leg}
\end{figure}
\begin{figure}[t]
    \centering
    \includegraphics[width=0.99\linewidth]{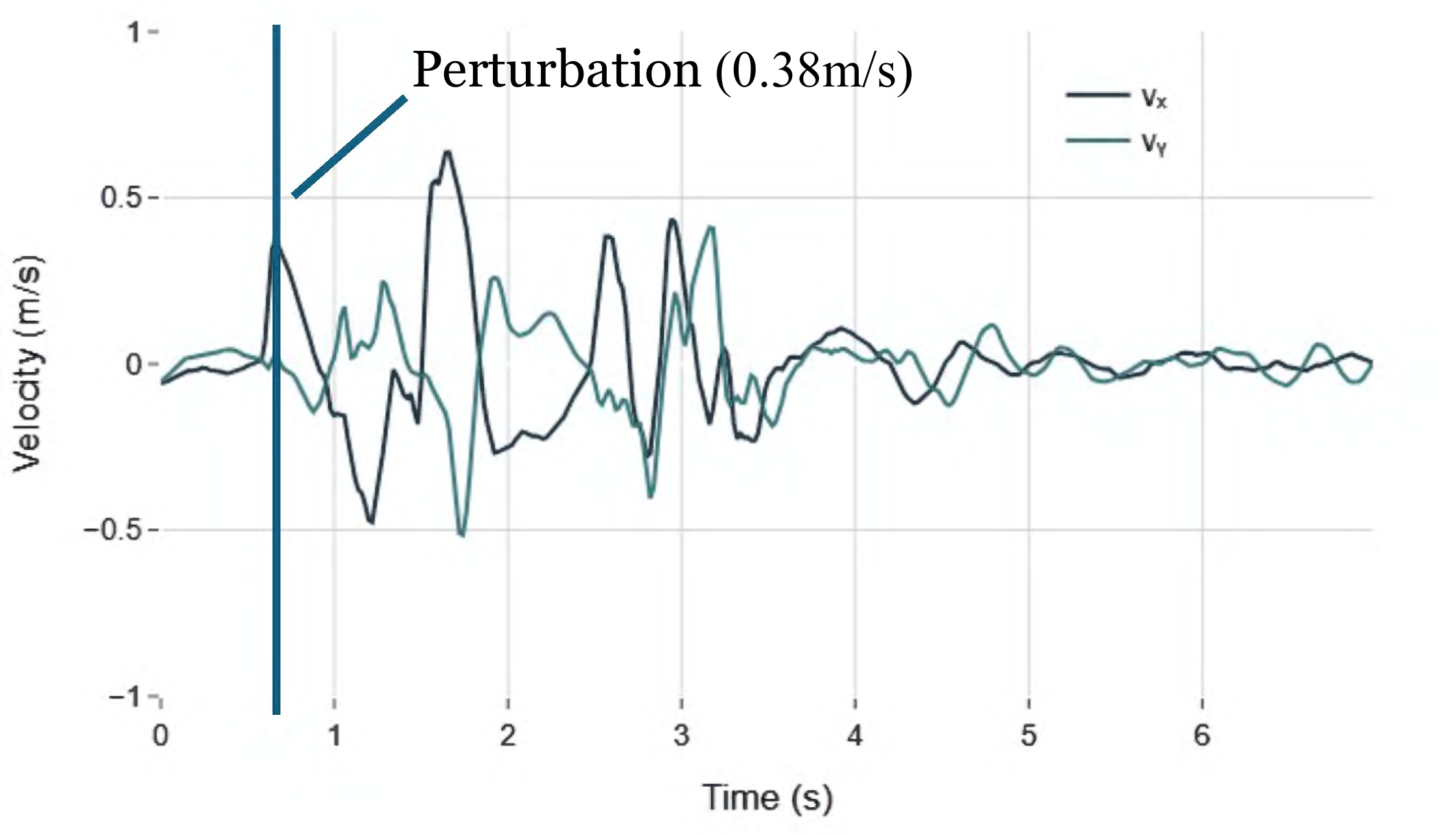}
    \caption{Perturbation disturbance rejection from standstill. The graph shows the linear velocity odometry over time. The robot was pulled forward ($x$-axis direction) at $\SI{0.38}{m/s}$ at $t=\SI{0.66}{s}$. The robot took steps to stabilize the torso orientation. }
    \label{fig:push}
\end{figure}

\subsection{Simulation Efficiency for Parallel Mechanisms}

A primary concern with incorporating closed-chain mechanisms into GPU-accelerated simulators is computational overhead. 
To quantify this, we benchmarked several BRUCE simulation variants, including a simplified serial chain model and models that incrementally incorporated the 4-bar, 5-bar, and differential closed-chain constraints. 
Each model executed 400 time steps with 8,192 environments, on an Intel i9-13900K CPU and Nvidia RTX4090 GPU.

As summarized in \tab{tb:performance}, the inclusion of all three parallel mechanism constraints introduced only a $3.4$ \% increase in per-step simulation time relative to the unconstrained model. Differential constraints contributed the most significant overhead, consistent with Section~\ref{sec:differential} due to two additional equality constraints, while the 4-bar constraint's impact was negligible. The Just-In-Time (JIT) compilation time increased $10.5$ \% with all parallel mechanisms constraints. For typical reinforcement learning workloads, this one-time initialization cost is amortized and insignificant compared to the overall training duration.

\begin{table}[t]
\centering
\caption{Relative simulation overhead for each parallel mechanism on BRUCE. Measurements from MJX with 8,192 parallel environments over 400 time steps each.}
\begin{tabular}{lccccc}
\toprule
 & Simplified & 4-Bar & 5-Bar & Differential & All \\
\midrule
Time/step  & 0.52 [µs] & \textbf{0.0\%} & \textbf{+1.2\%} & \textbf{+2.3\%} & \textbf{+3.4\%} \\
Steps/sec & 1,907,040 & \textbf{0.0\%} & \textbf{-1.2\%} & \textbf{-2.3\%} & \textbf{-3.4\%} \\
JIT time & 16.07 [s] &  \textbf{0.0\%} & \textbf{+4.3\%} & \textbf{+6.9\%} & \textbf{+10.5\%} \\
\bottomrule
\label{tb:performance}
\end{tabular}
\end{table}


\subsection{Sim-to-Real Transfer: Policy Validation on Hardware}

To validate our approach, we deployed policies trained with high-fidelity closed-chain constraints directly on a BRUCE hardware unit, without additional fine-tuning.

\subsubsection{Experimental Setup}
BRUCE's onboard computer (Khadas Edge2, 8-core 2.25 GHz Cortex-A76) executes all inference and control. The RL policy runs at $\SI{50}{Hz}$, whereas the convex MPC baseline \cite{bruce} is at $\SI{500}{Hz}$. Control input in both MPC and RL cases consists of desired body linear and angular yaw velocities.
We evaluate performance on a variety of real-world surfaces: foam, synthetic grass, and smooth concrete, such as in \fig{fig:locomotion_smooth}.

\subsubsection{Locomotion Robustness and Adaptability}
The RL policy demonstrates an adaptive walking style, standing still when no velocity commands are provided and dynamically transitioning to a natural gait once a disturbance is felt or commanded velocities are received.

\paragraph{Standstill stability} BRUCE RL policy has demonstrated an agile response to various perturbations. \fig{fig:one_leg} shows the RL policy reacting to a sideways disturbance with only the right leg and the left leg was in the air for $1.07$ s, which is more natural, human-like sideways balancing behavior. 
\fig{fig:push} graphs the perturbation rejection from standstill, which is more challenging since the robot has to quickly react to the disturbance and take a step if necessary. At $t=\SI{0.66}{s}$, a perturbation of $\SI{0.38}{m/s}$ in the $+\dot{x}$ (forward) direction was applied, which is significant given the scale of the robot. The RL policy first attempted to pull back, then took two steps forward as seen in the linear velocity odometry in \fig{fig:push}. The robot torso reached steady state after $\SI{5}{s}$ from the push. This showcases RL policy adaptiveness and stability.

\paragraph{Walking} In \fig{fig:locomotion_smooth}, the RL policy was set to step at $\SI{1.9}{Hz}$, and the measured gait frequency was $\SI{1.91}{Hz}$ over $\SI{10}{s}$ of straight walk. However, the RL policy was able to adjust the gait frequency and sequence naturally in response to external disturbances.  
In contrast, the MPC controller struggles to respond to disturbances from a standstill, requiring constant stepping. Although the gait phase time is adaptable in MPC, the sequence is fixed.

\subsubsection{Surface Generalization}
\tab{tab:surface_success} summarizes the performance of RL and MPC controllers for in-place stepping across different surfaces. 
The RL policy consistently succeeded on all tested surfaces, including highly resistive synthetic grass, high-friction foam, and smooth, low-friction concrete. 
In contrast, the MPC controller could sustain stepping on synthetic grass for $15$ minutes but failed to operate reliably on foam and concrete despite tuning efforts. 
For our RL policy, it failed due to excessive bouncing of the foot when stepping, which is more commonly observed on hard concrete. In the future, such contact impedance should be better modeled and domain-randomized in the training.

\begin{table}[t]
  \centering
  \caption{In-place stepping success across surfaces}
  \label{tab:surface_success}
  \begin{tabular}{lccc}
    \toprule
    Method        & Synthetic Grass & Foam & Smooth Concrete \\
    \midrule
    RL Policy     & \checkmark      & \checkmark & \checkmark \\
    MPC Baseline  & \checkmark      &            &             \\
    \bottomrule
  \end{tabular}
\end{table}

\subsubsection{Locomotion Speed}
On hardware, the RL policy achieved a peak forward walking speed of $\SI{0.18}{m/s}$ on a flat, smooth concrete floor. 
While the maximum speed is $28$\% lower than the reported MPC maximum of $\SI{0.25}{m/s}$ \cite{BRUCE_MPC}, the RL policy can operate on a broader range of surfaces.

\subsubsection{Policy Inference and Computation Cost}
The policy runs at $\SI{50}{Hz}$, and the policy network requires $\SI{3.2}{ms}$ per inference. 
The MPC pipeline computes faster but must run at a higher frequency of $\SI{500}{Hz}$, perform online quadratic programming, and requires separate kinematics computations. Our policy architecture, enabled by the closed-chain simulation, eliminates the need for extra forward and inverse kinematics or state estimation on hardware, reducing CPU consumption and system complexity. 

\subsubsection{Ablation Study: Effect of Parallel Mechanism Fidelity on Sim-to-Real Gap}
We compared sim-to-real performance with and without explicit simulation of passive joint compliance and backlash. Policies trained without such a parallel mechanism's backlash modeling were brittle in terms of hardware compliance. 
They tended to fail during backward tipping events, as BRUCE's heel is short and passive compliance limits force exertion at the heel. 
Including this compliance and backlash in the simulation led to qualitatively improved disturbance response strategies, with earlier footlift and stepping actions.

\section{Discussion and Limitation}

\subsection{Benefits in RL Training}
\subsubsection{System identification} The motor system identification is a crucial part of zero-shot RL policy deployment. However, system identification on parallel mechanisms is challenging since the passive joints often lack sensors. While Jacobian stiffness modeling \cite{tanaka2023scaler} can provide estimates, integrating them into the simulation is challenging. Simulating the parallel mechanisms simplifies the system identification process as the same motors drive the lower body.

\subsubsection{Torque and Velocity} Incorporating parallel mechanisms directly into the simulation is a critical step toward achieving a truly end-to-end RL policy. Aligning the simulation’s action space with the hardware’s actuation space enables a more seamless policy transfer. This closed-chain simulation allows for non-positional commands, such as torque or velocity, without requiring explicit inverse dynamics or Jacobian computations.

Even when using joint position as the control input, as done in this work, the reward function includes an energy minimization term in Section~\ref{sec:reward}. This term becomes inaccurate if the mechanism model is simplified or a kinematic joint space is used. 
For example, in general gear ratio configurations of differential pulleys, minimizing output torques does not guarantee a reduction in actuator torques.

\subsection{Nonlinear Actuation Joint Limit}
A key limitation in simulating closed-chain kinematics is the nontrivial and configuration-dependent nature of joint limits in actuated coordinates. This issue can be mitigated in simulation by setting passive joint limits or adding self-collision constraints to enforce physical realism. In hardware, however, the lack of sensing at passive joints makes it nontrivial to enforce these limits without forward kinematics. 
For actuators driving parallel mechanisms, the action scale may require additional consideration when the position is a policy action. 
In our setup, the RL policy learned kinematic joint limits implicitly through training, and we have not encountered any joint limit violations in hardware deployment.

\subsection{Parallel Mechanisms with Compliant Members}
While the current simulation models only include backlash effects, compliant parallel mechanisms—such as those with embedded springs—are common in legged systems \cite{shin2019_five_bar}. The equality constraint formulation in \eqref{eq:mujoco_const} can be tuned to approximate spring-damper behavior for small displacements. However, modeling large-deformation compliant elements or flexure members would require additional extensions that are not addressed in this work.

Beyond enabling torque- and velocity-space control, our closed-chain simulation faithfully captures actuator-level torque mappings and kinematic singularities (e.g., in the four-bar and five-bar linkages), which we view as a key benefit for policy learning and evaluation. Although we observe qualitative gains from closed-chain modeling, a controlled RL comparison against simplified serial/kinematic models under identical training and domain-randomization settings is deferred to future work to quantify the effect more precisely.

\section{Conclusion}
This work presents general formulations of three types of parallel mechanisms for 
an end-to-end curriculum reinforcement learning. Incorporating full closed-chain kinematic constraints in MJX simulation enables learning directly in the hardware actuator space, preserving the mechanical intelligence of parallel actuation, the nonlinear linkage transmission, and the singularity. 
We demonstrated our approach on BRUCE with differential pulleys, 5-bar, and 4-bar linkages. 
Our experiments demonstrate that an entirely end-to-end policy, trained using our method and seamless sim-to-real transfer, achieves improved performance compared to MPC baseline. 
Comparisons with model predictive control (MPC) further validate the effectiveness of our RL policy in real-world deployment. Our closed-chain model preserves the true actuator-to-output torque relationships and makes the four- and five-bar singularities explicit, yielding richer signals for policy training and evaluation. This study highlights the importance of embracing mechanical structure in learning-based control and opens the door for broader integration of parallel mechanisms in legged robot RL training pipelines.


\bibliographystyle{IEEEtran}
\bibliography{main}


\end{document}